\documentclass[12pt,letterpaper]{article}
\newcommand{\Author}{Michael Stewart}
\newcommand{\Title}{Natural Language Feature Selection via Cooccurrence}

\title{\Title}
\date{\today}
\author{\Author}

\usepackage[utf8]{inputenc}
\usepackage{graphicx, epstopdf, wrapfig, float}    
\usepackage{alltt, amsmath, multirow, tabularx, ragged2e, booktabs, caption, mdwlist, dcolumn}
\usepackage{amsmath,amsfonts}  
\usepackage{sidecap}                           

\usepackage{natbib}                            
\renewcommand{\cite}[1]{[\citealt{#1}]}        

\usepackage{hyperref}                        
\usepackage[usenames,dvipsnames]{xcolor}     
\hypersetup{
    pdfauthor={\Author},    
    pdftitle={\Title},      
    unicode=false,          
    pdftoolbar=true,        
    pdfmenubar=true,        
    pdffitwindow=false,     
    pdfstartview={FitH},    
    colorlinks=true,        
    linkcolor=black,        
    citecolor=black,        
    filecolor=magenta,      
    urlcolor=Blue,          
    }
\usepackage[compact]{titlesec} 

    \usepackage{fancyhdr}
    \usepackage{layout}     
\everymath{\displaystyle}    

\begin{document}
\maketitle

\begin{abstract}
Specificity is important for extracting collocations, keyphrases, multi-word and index terms \cite{Newman2012}. It is also useful for tagging, ontology construction \cite{Ryu2006}, and automatic summarization of documents \cite{Louis2011,chali2012}. Term frequency and inverse-document frequency (TF-IDF) are typically used to do this, but fail to take advantage of the semantic relationships between terms \cite{Church1995}. The result is that general idiomatic terms are mistaken for specific terms.
We demonstrate use of relational data for estimation of term specificity. The specificity of a term can be learned from its distribution of relations with other terms. This technique is useful for identifying relevant words or terms for other natural language processing tasks.
\end{abstract}

\section{Motivation}
A deeper understanding of the semantics in natural language can help overcome limitations of basic statistical methods that lack it. One fundamental property of natural language tokens is specificity. Specificity was defined for a term as the number of documents to which the term pertains \cite{Jones1972}, but has since become more abstract in order to apply to multi-word terms and other natural language tasks \cite{Frantzi1998}. The common definition is that a ``specific'' term has meaning within a relative subdomain, while a ``general'' term may apply to entire domains of study.

Specificity itself is usually estimated with frequency statistics such as term-frequence and inverse-document frequency (TF-IDF). We will try to do better than TF-IDF by taking advantage of the underlying semantic relationships between terms.
Our primary assumption is that two terms which are strongly related will tend to occur together. This is known as the \emph{latent relation hypothesis} \cite{akbik2012}. We will use this to infer relations through term collocations. We connect concept relations to a notion of specificity using the theory of a ``semantic hierarchy'' \cite{Chodorow1985}. In a semantic hierarchy, ``high level'' terms are general, and exist above ``low level'' terms, and a high level term is connected to specific terms which are related to it.

\begin{figure}[h]
    \centering
    \includegraphics[trim= 10px 10px 10px 20px,width = 1.0\textwidth]{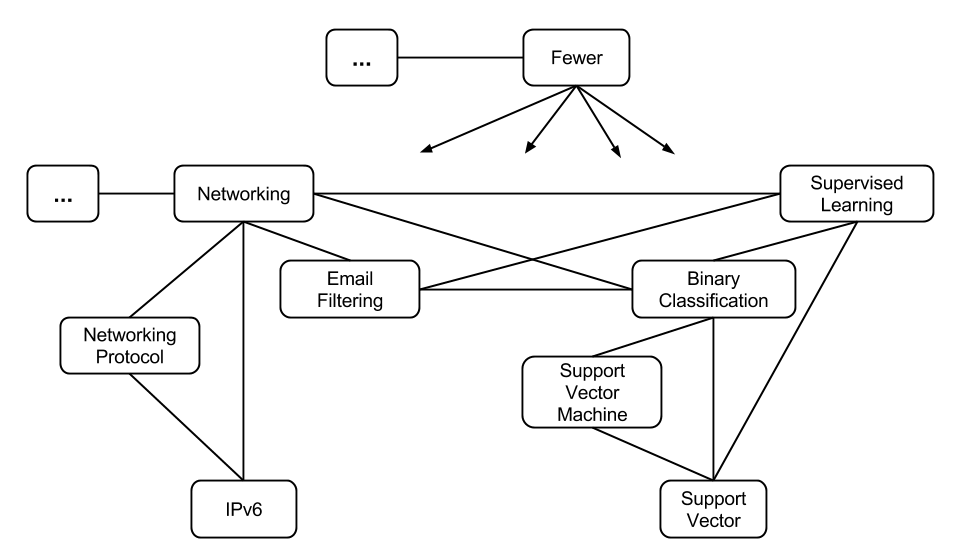}
    \caption{A hand-crafted example of a semantic hierarchy. High level concepts have many weak relationships. Low level concepts have few strong relationships. }
    \label{fig:semanticHierarchyExample}
\end{figure}

We will create a new method of inferring specificity by using a simple cooccurrence model of relations together with the idea of a semantic hierarchy. This method is based on the semantics of the terms involved, and so robust against functional words which TF-IDF tends to fail at. This results in the high-precision selection of terms appropriate for tasks such as tagging where a relatively small number of terms are desired.


\section{Prior Work}
Prior work on term specificity has used frequency statistics such as TF-IDF \cite{Church1995}, context measures such as C/NC-value \cite{Frantzi1998,Caraballo1999,Ryu2006}, and latent-space-analysis techniques based on term informativeness \cite{Hogan2007,Kireyev2009}.

Relationship extraction between terms follows two strategies: the use of statistical measures \cite{He1999,Ryu2006,Hogan2007} and pattern-based information extraction techniques \cite{Navigli2004,akbik2012}.

Statistical measures attempt to capture a condional probability such as ``given term A has occurred, what is the probability of term B occurring?'' \cite{He1999} and more primitive frequencies such as cooccurrence and individual term frequency \cite{Ryu2006}. They include bag-of-words frequencies as well as contextual measures. Context methods examine the distribution of modifier terms which immediately precede or follow the term in question. \citet{Ryu2006} describe the semantic model behind context methods: ``Distribution of adjective-term relation refers to the idea that specific nouns are rarely modified, while general nouns are frequently modified in text.'' These unsupervised techniques are more popular for large-scale use on the web because they require less training and fine-tuning than the pattern-based methods \cite{Rosenfeld2007,akbik2012}.

Pattern-based information extraction relies on specialized parsers to extract ternary relations: two operand terms and a third relation operator. e.g. For the sentence ``SVMs are a kind of binary classification.'' a parser might identify the two terms ``SVM'' and ``binary classification'' as well as the ``is a'' relation between them. This parsing technique relies heavily on prior syntactic knowledge and has problems identifying relations that aren't represented in a single sentence \cite{Navigli2004,akbik2012}. Performance is dependent on pre-processing methods and the parsing language pattern used \cite{Etzioni2011, Gupta2011}.

Our contribution is the combination of a relatively fast relation extraction technique with the semantic model for specificity inference.


\section{Models}
\subsection{Relation Extraction}
Since this relational model is essentially a knowledge representation used by later techniques, we choose a crude and computationally fast method for inference of term relations. Via the \emph{latent relation hypothesis} mentioned earlier, we assume that a semantic relationship exists between two terms based on cooccurrence frequency. Relations between concepts will only be modeled implicitly.

The corpus is discretized into observation units. Units could be defined by sentence breaks, paragraph breaks, document breaks, etc. To collect occurrence statistics, each unit is treated as an independent observation of terms. e.g. If we used sentence breaks to define observation units, the sentence ``SVMs are a kind of binary classification.'' would imply a stronger relation between ``SVM'' and ``binary classification''.

Compared to parsing methods for relation extraction, this method is not constrained to sentence-level analysis and can be done more easily. However, it doesn't learn the type or quality of the relation.
Where parsing techniques treat relations as either present or not-present, cooccurrence will produce relations between every term. Fortunately, these relations are identified by statistics which can be interpreted as confidence or \emph{strenth} of the relation. A strong relation is one that a parser might identify e.g. the ``is a'' relation between two terms in a sentence. A weak relation is one that is semantically trivial and unlikely for a parser to find e.g. ``they are different measures used in different techniques but occur in the same corpus'' is a relation which will probably not be explicit in natural text, but would result in a weak cooccurrence relation.

The assumption that cooccurrence implies a relationship is not always correct. The result of our method is a full graph which requires some sort of pruning to avoid false relations. For our applications, we do not prune the graph to test how useful it is in this state.

Identifying cooccurrence statistics requires $O(n^2)$ work for vocabulary of size $n$.

\subsection{Specificity}
Once we have an idea of relations between terms, we use a semantic model to infer specificity. We assume a semantic hierarchy exists between all terms. It is not a strict tree hierarchy in that lower terms may be connected to multiple parents. However, it is still tree-like in that a more specific term will have fewer relations with other terms because it is semantically unrelated to terms outside its domain. A more general term will have a larger number of relations with specific terms within its domain as well as relations with peers and parents. At the top of the hierarchy are very general terms which serve functional purposes in writing or are not domain-specific. We do not have to explicitly model the hierarchy, but will infer the specificity of a term by its distribution of relations with other terms.

\begin{figure}[h]
    \centering
    \includegraphics[trim= 00px 00px 00px 00px,width = 1.0\textwidth]{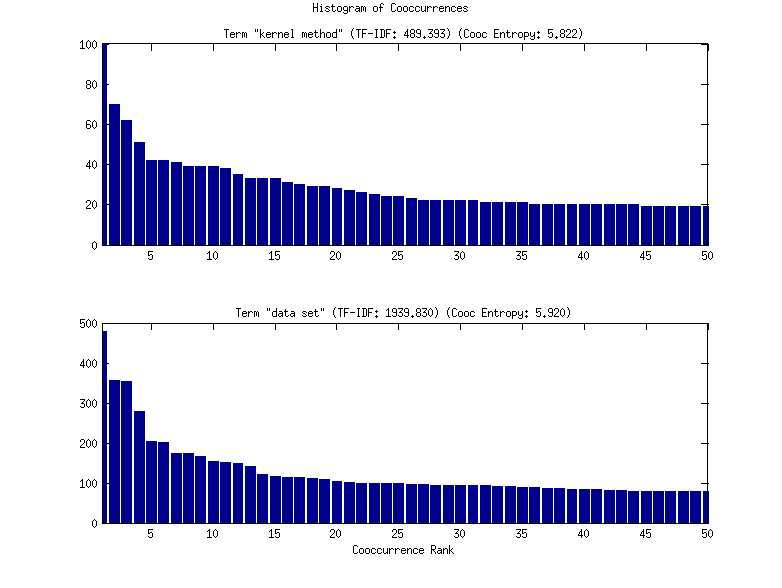}
    \caption{Histograms of cooccurrence between one term with all other terms, sorted by coocurrence rank. Hand-picked examples for terms ``kernel method'' and ``data set''. The steepness of the curve reflects the specificity of the term.}
    \label{fig:histogramExample}
\end{figure}

Figure \ref{fig:histogramExample} shows an example of the very general word ``data set'' which has higher TF-IDF than the specific word ``kernel method''. TF-IDF would consider ``data set'' a more relevant term than ``kernel method''. A specific word like ``kernel method'' will have a restricted domain of relations with other terms, so its relations are distributed in a more predictable manner. The result is that its relation distribution has a lower Shannon entropy than the general word ``data set''.

We will use the normalized version of this cooccurrence distribution to approximate the probability that term $t$ has a relation with term $x$. 
The cooccurrence approximation for a relation between term $t$ and term $x_i$ will be: 
$$p_c^t(x_i) = \frac{ M_{t,x} }{ \sum_{i=1}^N M_{t,i} } = \frac{\text{count of cooccurrences between $t$ and $x_i$}}{\text{count of all cooccurrences with $t$}}$$
Where $N$ is the vocabulary size and $M$ is a matrix of nonnegative cooccurrence counts. $M_{a,b}$ is the total number of observation units which contain both term $a$ and term $b$.
The Shannon entropy formula is useful because it captures the idea of predictability and we expect specific terms to have more predictable coocurrence distributions. 
$$H^t(X) = - \sum_i^N p_c^t(x_i) \log \left( p_c^t(x_i) \right) $$
Terms that have specific relation patterns receive a lower score. The results are ranked from low-to-high to produce the most specific terms first. 

Altogether, there is $O(n^2)$ work required to compute this value for all terms in vocabulary of size $n$. Since TF-IDF is $O(n)$, our method is more costly than computing TF-IDF from the same data. 

\section{Methods and Results}

\subsection{Application in Keyphrase Extraction}
Index term and keyphrase extraction is a task which requires extraction of ``main terms and concepts in a document'' \cite{Newman2012}. This is analogous to extracting both low-level terms as well as terms which are general but still within the domain of a particular document. That is, we no longer want to identify the most specific terms, but a particular range of terms. With this aim, we use alternate versions of the probability estimate $p_c$ based not directly on cooccurrence. The alternate estimators are $p_v$ and $p_{mi}$, estimates based on covariance and mutual information.

\begin{figure}[H]
    \centering
    \label{fig:keyphraseRegimes}
    \includegraphics[width = 0.8\textwidth]{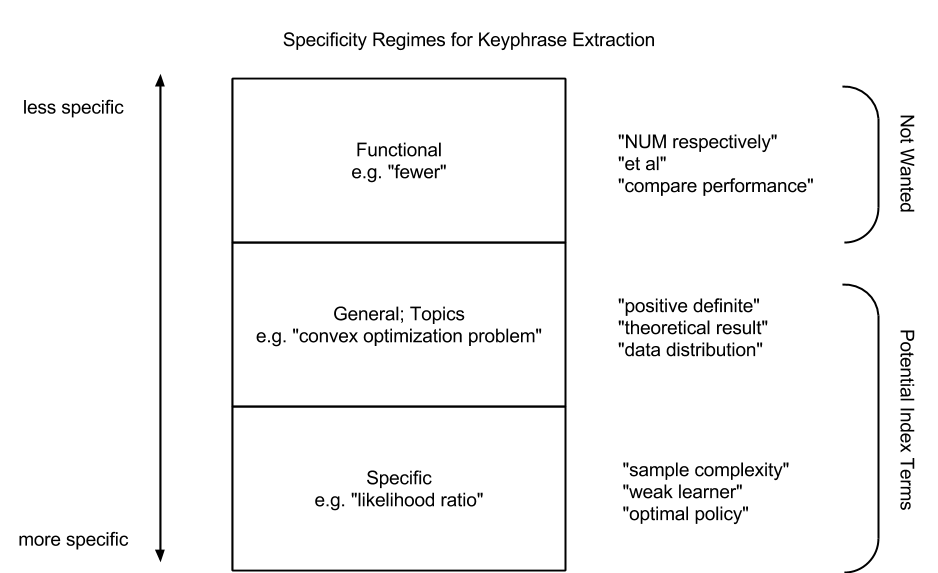}
    \caption{Sample terms taken from the median and 95th percentiles of our cooccurrence method.}
\end{figure}

Our primary data set was a set of machine learning and neurobiology articles from NIPS. It has 132262 unigrams, or 86529 terms after using a segmenter \cite{Newman2012}. The segmented term results are shown below, but similar results were found with unigrams and alternate data sets including a set of ACL research article texts and PubMed search results.

The covariance between terms was calculated from observation data. Positive covariances are assumed to represent the strength of a relationship between terms, so non-positive covariances are discarded. A specific term will have most of its relationship mass concentrated in a smaller number of other terms, so an entropy calculation is taken over the positive-covariance distribution for each term. This method treats negative correlations as zero and so ignores the negative relationships between terms of different domains.

A second measure uses mutual information in an attempt to leverage the negative relations ignored before. Data is binarized to represent ``present'' or ``not present'' for each term in each observation. Mutual information is calculated by 
$$I(X;Y) = \sum_{y \in Y} \sum_{x \in X} 
                 p(x,y) \log{ \left(\frac{p(x,y)}{p(x)\,p(y)} 
\right) }$$
In a binarized frequency format, there are four cases which contribute to the mutual information calculation: two where a single term is present, one where both are present, and one where neither is present. A specific term will tend to have high mutual information with many others. Due to the many terms outside its domain, a specific term tends toward an XOR pattern with others: where one is present, the other is not. This results in higher mutual information between these terms. Additionally, the specific term will have high mutual information with terms in its domain. On the other hand, general terms which are still domain-specific gain mutual information from their descendents in the semantic hierarchy. It is the functional terms which we want to filter out which will have lower mutual information in more cases. When an entropy estimate is calculated over this distribution with other terms, the results are ranked high-to-low to produce domain-specific terms first.

\begin{figure}[p]
\begin{center}
    \includegraphics[width=0.8\textwidth, trim= 0px 30px 0px 20px]{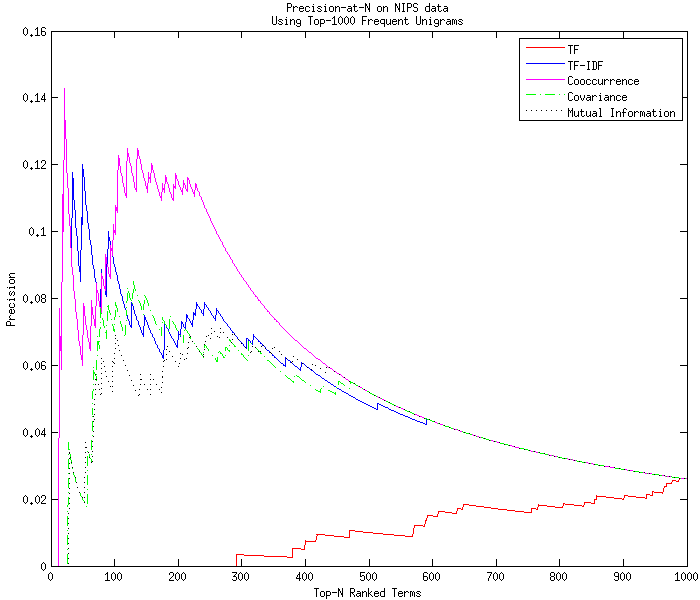}
\end{center}
\end{figure}

\begin{figure}[p]
\begin{center}
    \includegraphics[width=0.8\textwidth, trim= 0px 30px 0px 20px]{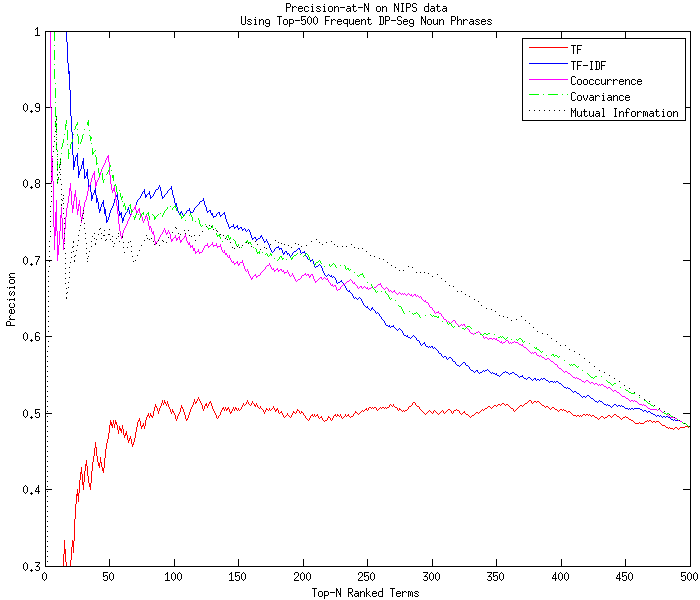}
\end{center}
\end{figure}

We used document-breaks to define observation units so that each method used the same input data: bag-of-words frequency counts for each document. When used to rank terms directly, the cooccurrence methods appear comparable to TF-IDF at this task. They have very similar precision scores for the top-N ranked terms compared to TF-IDF.

\begin{figure}[p]
\begin{center}
    \includegraphics[width=1.0\textwidth, trim= 60px 20px 60px 0px]{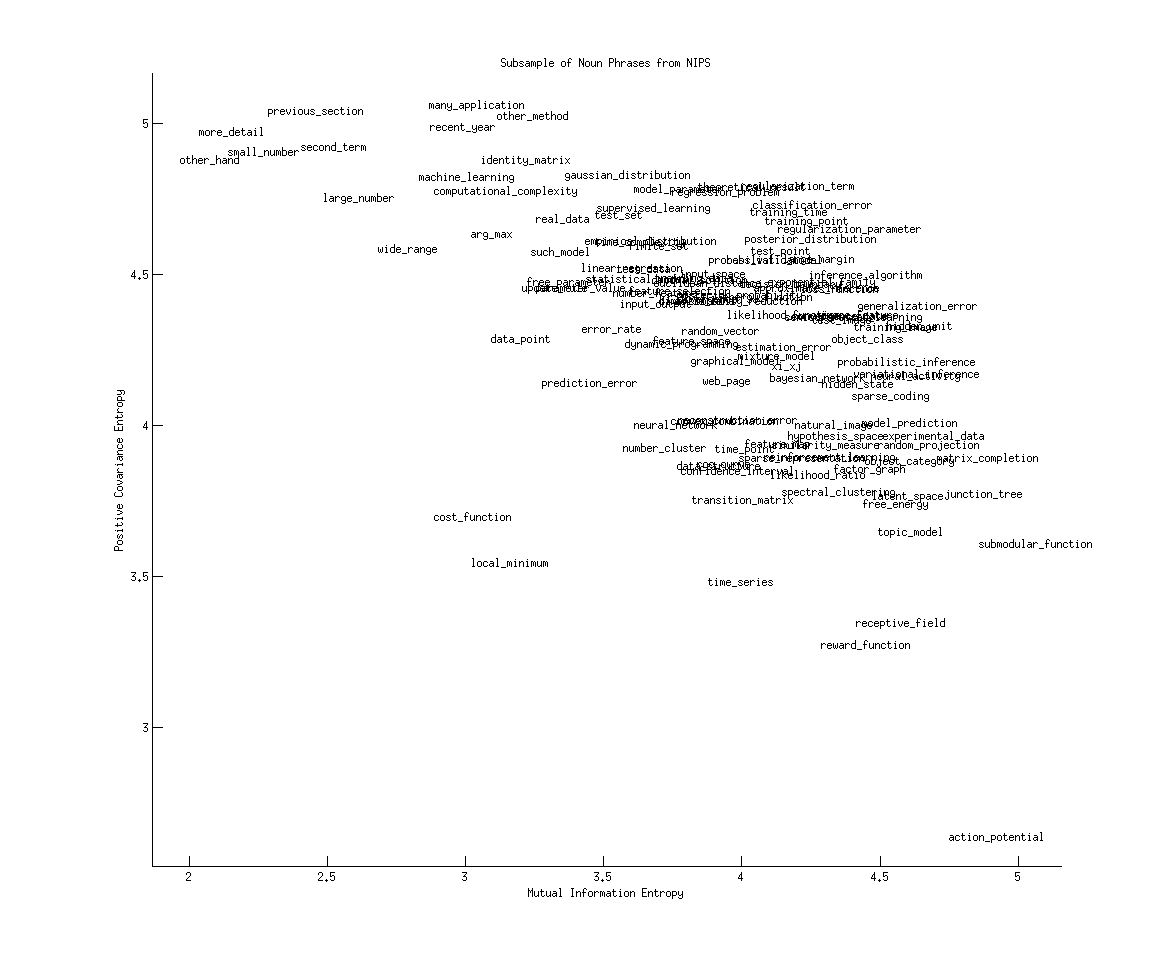}
    \caption{From this plot of M.I. and Covariance, M.I. appears to do a better job of separating out the functional terms from index terms. For example, a support-vector machine trained only on this entropy of mutual information will use a decision bounary at the 3.5 value. Training on entropy of covariance alone places the boundary just under 4.5 and results in a high false-negative rate as general terms are excluded. }
\end{center}
\end{figure}

For each of these top 4 measures, we performed 30-fold cross validation over 500 terms using single-feature support-vector machine classifiers. Classifiers did best when trained on entropy of mutual information. Performance increases with the amount of work required to compute the feature.

\begin{figure}[H]
\begin{center}
    \includegraphics[width=1.0\textwidth, trim= 60px 20px 60px 0px]{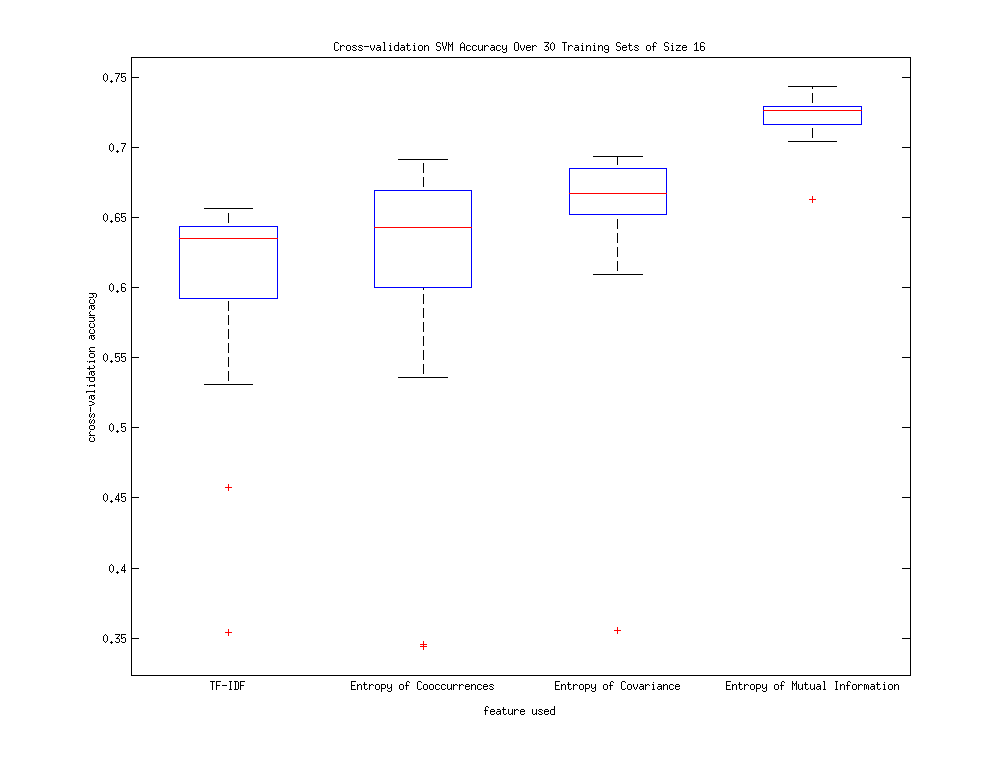}
\end{center}
\end{figure}

\begin{table}[p]
\begin{center}
\caption{ The top 20 terms by the relative increase in ranking compared to the cooccurrence method. }
\begin{tabular}{l c c}
Term                        & Increase Under TF-IDF     & Golden \\ \hline
data point                     	& 482 	& 1	\\
section NUM                    	& 481 	& 0	\\
et al                          	& 481 	& 0	\\
equation NUM                   	& 476 	& 0	\\
figure NUM                     	& 468 	& 0	\\
training data                  	& 449 	& 1	\\
NUM figure NUM                 	& 437 	& 0	\\
other hand                     	& 435 	& 0	\\
table NUM                      	& 432 	& 0	\\
show how                       	& 421 	& 0	\\
\end{tabular}
\end{center}
\end{table}

\begin{table}[p]
\begin{center}
\caption{ The top 20 terms by the relative increase in ranking compared to TF-IDF. }
\begin{tabular}{l c c}
Term                        & Increase Under Cooccurence  & Golden \\ \hline
model predict                  	& 405 	& 0	\\
generative process             	& 397 	& 1	\\
follow lemma                   	& 369 	& 0	\\
sufficient condition           	& 359 	& 1	\\
model prediction               	& 354 	& 1	\\
kernel method                  	& 342 	& 1	\\
online learning                	& 338 	& 1	\\
convex function                	& 331 	& 1	\\
dash line                      	& 331 	& 0	\\
proof theorem NUM              	& 331 	& 0	\\
convex optimization problem    	& 330 	& 1	\\
\end{tabular}
\end{center}
\end{table}


\section{Future Work}

\subsection{Evalutation of Specificity Model}
The specificity model could be evaluated directly by comparison with existing human-curated ontologies such as the Medical Subject Headings (MeSH) provided with NCBI publications.

\subsection{Application in Automatic Summarization}
Recent work shows that specificity measures can aid in generating better summaries of large texts \cite{Louis2011,chali2012}. To test generality of the specificity measure, it could be implemented with a summarization technique like this. 

\subsection{Structure Learning for Semantic Hierarchy Models}
The semantic hierarchy which was only used implicitly before could be modeled explicitly.

When plotting terms based on the covariance and mutual information entropies, the results from above emerge. General terms tend to occur in the upper regions of the plot, and very general stop-word terms tend to occur in the left side:

\begin{figure}[H]
\begin{center}
    \includegraphics[width=1.1\textwidth, trim= 60px 20px 60px 0px]{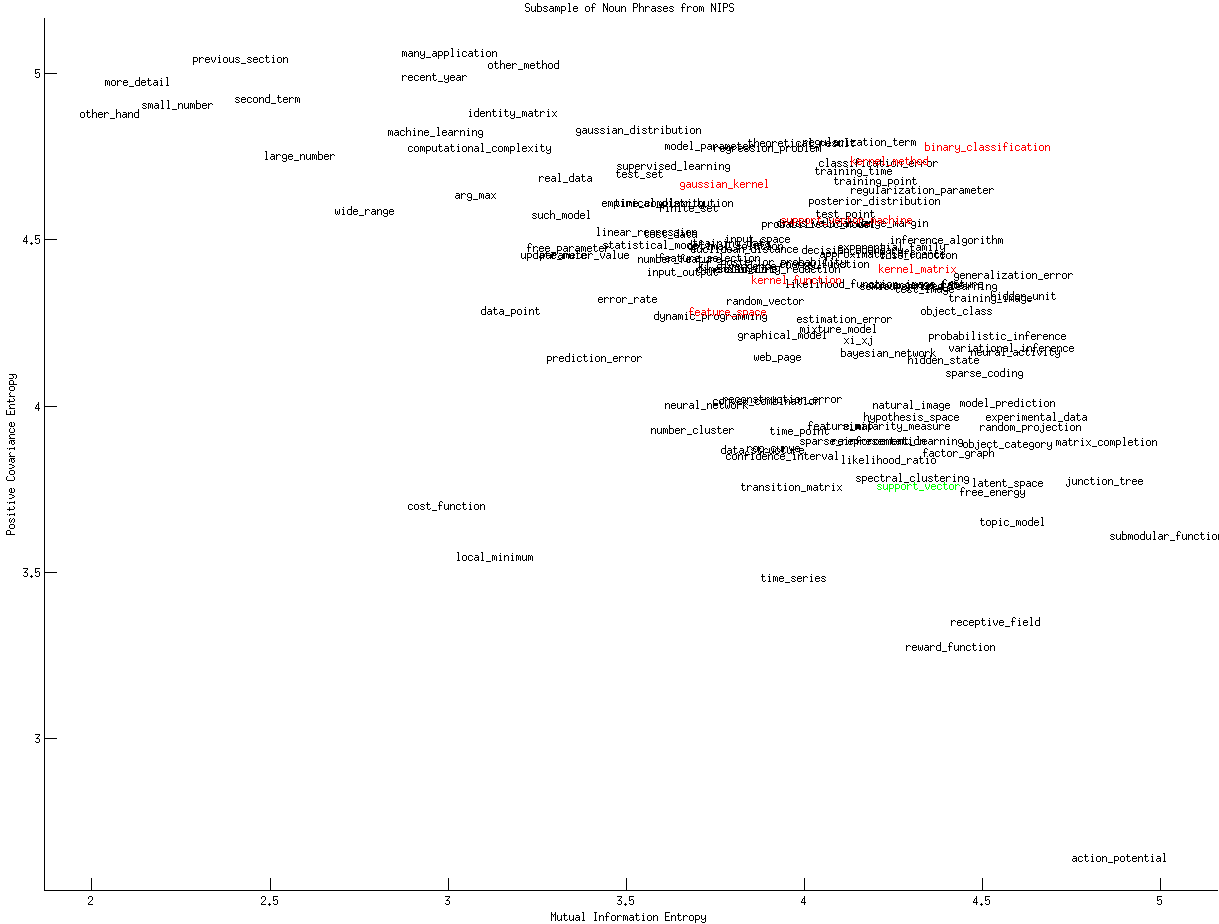}
    \caption{Anecdotal hierarchy example}
\end{center}
\end{figure}

The specific term ``support vector'' is highlighted in green. Top related terms to it are highlighted in red. These are the terms which had the highest mutual information with ``support vector''. Other highly related terms to ``support vector'' which I didn't plot were ``training point'', ``training set'', and ``training data''. I didn't plot them because they overlap and are difficult to see. These ``training X'' terms tend to be associated with most supervised learning techniques, and are considered less specific than ``support vector''. As a low-level specific term, our model doesn't expect it to be related to many other low-level terms.

\bibliographystyle{plainnat}
\bibliography{bib.bib}

\end{document}